\documentclass{article}

\usepackage[preprint]{neurips_2024}

\usepackage[utf8]{inputenc}
\usepackage[T1]{fontenc}
\usepackage{amsmath,amssymb}
\usepackage{graphicx}
\usepackage{hyperref}
\usepackage{booktabs}
\usepackage{multirow}
\usepackage[inline]{enumitem}

\newcommand{\Var}{\operatorname{Var}}

\title{Localising Dropout Variance in Twin Networks}
\author{
  Cooper Doyle\thanks{Correspondence to: cooper.doyle@cba.com.au} \\
  Commonwealth Bank of Australia \\
}
\date{July 4, 2025}

\begin{document}
\maketitle

\begin{abstract}
Accurate individual treatment-effect estimation demands not only reliable point predictions but also uncertainty measures that help practitioners \emph{locate} the source of model failure. We introduce a layer-wise variance decomposition for deep twin-network models: by toggling Monte Carlo Dropout independently in the shared encoder and the outcome heads, we split total predictive variance into an \emph{encoder component} ($\sigma_{\mathrm{enc}}^2$) and a \emph{head component} ($\sigma_{\mathrm{head}}^2$), with $\sigma_{\mathrm{enc}}^2 + \sigma_{\mathrm{head}}^2 \approx \sigma_{\mathrm{tot}}^2$ by the law of total variance. Across three synthetic covariate-shift regimes, the encoder component dominates under distributional shift ($\rho_{\mathrm{enc}}=0.53$) while the head component becomes informative only once encoder uncertainty is controlled. On a real-world twins cohort with induced multivariate shift, only $\sigma_{\mathrm{enc}}^2$ spikes on out-of-distribution samples and becomes the primary error predictor ($\rho_{\mathrm{enc}}\!\approx\!0.89$), while $\sigma_{\mathrm{head}}^2$ remains flat. The decomposition adds negligible cost over standard MC Dropout and provides a practical diagnostic for deciding whether to collect more diverse covariates or more outcome data.
\end{abstract}

\section{Introduction}
\label{sec:intro}

Deep twin-network models such as TARNet~\cite{Shalit2017} and DragonNet~\cite{Shi2019} have become standard tools for estimating individual treatment effects (ITE) from observational data. These architectures share a covariate encoder across treatment arms and branch into separate outcome heads, mitigating selection bias through representation learning. Monte Carlo Dropout~\cite{Gal2016} and deep ensembles~\cite{Lakshminarayanan2017} can equip such models with uncertainty estimates, but these estimates are monolithic: a single variance score that does not distinguish whether the model is uncertain because its encoder has poor coverage of the covariate space or because its outcome heads lack sufficient signal in a well-covered region.

This distinction matters in practice. If encoder-level variance dominates, the model is extrapolating---the appropriate response is to collect data from underrepresented covariate regions. If head-level variance dominates, the encoder has adequate coverage but the outcome mapping is noisy---the appropriate response is to gather more outcome observations in the existing covariate distribution, or to increase head capacity. A single uncertainty score conflates these two failure modes and provides no guidance on which corrective action to take.

We propose a simple, architecture-aware decomposition that separates these signals. The key observation is that a twin network has a natural modular structure---encoder then heads---and the law of total variance provides an exact split of total predictive variance into two terms: one attributable to variation across encoder dropout masks (holding heads fixed) and one attributable to variation across head dropout masks (holding the encoder fixed). By running MC Dropout in three modes---encoder-only, heads-only, and both---we obtain the two components and verify their additivity, all at the cost of two additional sets of forward passes at inference time.

The decomposition provides a cheap, actionable diagnostic: empirically, the encoder component tracks covariate-shift-induced error far more reliably than total variance alone, and the head component becomes a strong error predictor precisely in the regime where encoder uncertainty is low.

Across three synthetic data generators with controlled covariate shift, we show that the encoder component dominates error prediction under shift ($\rho_{\mathrm{enc}}=0.53$ vs.\ $\rho_{\mathrm{tot}}=0.45$ in the strong-shift regime), while the head component is near-zero until high-encoder-uncertainty points are filtered out, at which point it rises above 0.5. On a real-world same-sex twins cohort with induced multivariate sampling bias, the encoder component spikes on out-of-distribution samples ($\Delta\sigma_{\mathrm{enc}}^2\!\approx\!0.004$) and becomes the primary error predictor ($\rho_{\mathrm{enc}}\!\approx\!0.89$), while the head component remains flat. These patterns suggest that the decomposition, despite its theoretical limitations, offers genuine diagnostic value for practitioners deploying treatment-effect models under distributional uncertainty.

\section{Related Work}
\label{sec:related}

\subsection{Uncertainty in Deep Learning}

Monte Carlo Dropout~\cite{Gal2016} interprets dropout at test time as approximate Bayesian inference, providing a scalable estimate of predictive uncertainty. Deep ensembles~\cite{Lakshminarayanan2017} average predictions from independently trained models to capture both predictive accuracy and uncertainty, often outperforming single-model Bayesian approximations. Deep Gaussian Processes~\cite{Damianou2013} offer a nonparametric alternative by stacking GP layers, though scalability remains a limitation. Recent work has revisited the foundations of uncertainty taxonomy, arguing that the aleatoric--epistemic boundary is less clean than commonly assumed and proposing richer categorizations~\cite{RethinkEU2412.20892, Wang2025}. Our work is sympathetic to this view: we focus on the architectural localisation of variance, which is well-defined regardless of how one taxonomises uncertainty.

\subsection{Causal Inference and Treatment Effect Estimation}

Heterogeneous treatment-effect estimation spans meta-learners (T-, S-, X-learner~\cite{Kunzel2019}), representation-learning architectures (TARNet~\cite{Shalit2017}, DragonNet~\cite{Shi2019}), and Bayesian nonparametric methods (BART~\cite{Chipman2010, Hill2011}, Causal Forests~\cite{Wager2018}). All of these can produce interval estimates, but none decomposes uncertainty by architectural module. A recent survey of deep causal models~\cite{Zhang2024DeepCausalSurvey} identifies the absence of structured, interpretable uncertainty quantification as an open problem. Our contribution addresses this gap by offering a practical module-level diagnostic that leverages the natural encoder--head structure of twin networks.

\subsection{Structured Uncertainty Decomposition}

The distinction between aleatoric and epistemic uncertainty is central to Bayesian deep learning~\cite{Kendall2017}. Multi-headed architectures that predict both mean and variance~\cite{Kendall2017}, and single-model methods for simultaneous estimation~\cite{NeurIPS2024UQSingleModel}, have made progress in computer vision and reinforcement learning. However, these methods decompose uncertainty by \emph{type} (data noise vs.\ model uncertainty), which requires strong assumptions about what each output captures. Our approach is orthogonal: we decompose by \emph{location} in the architecture (encoder vs.\ heads), which is unambiguous and requires no assumptions beyond the standard law-of-total-variance identity. The two approaches are complementary---a practitioner could in principle apply both---but we argue that the architectural decomposition is more directly actionable because it points to specific modules and, by extension, specific remedial strategies.

\section{Methodology}
\label{sec:method}

\subsection{Preliminaries and Notation}
Let $x\in\mathcal X\subseteq\mathbb R^p$ be a feature vector, $t\in\{0,1\}$ the binary treatment indicator, and $Y(t)$ the potential outcome under treatment~$t$.  We observe $Y = Y(T)$ for each unit.  A \emph{twin network} comprises
\begin{itemize}
  \item a shared encoder $\Phi(x;\theta_e)\colon\mathbb R^p\to\mathbb R^d$, and
  \item two outcome heads $f_0(z;\theta_0),f_1(z;\theta_1)\colon\mathbb R^d\to\mathbb R$,
\end{itemize}
so that the estimated individual treatment effect is
\[
  \hat\tau(x) \;=\; f_1\bigl(\Phi(x)\bigr)\;-\;f_0\bigl(\Phi(x)\bigr).
\]

\subsection{Monte Carlo Dropout}
Following Gal and Ghahramani~\cite{Gal2016}, we insert dropout after every hidden layer in both encoder and heads.  Each stochastic forward pass samples a binary mask, yielding
\[
  \hat Y_t^{(n)}(x)
  = f_t\bigl(\Phi(x;\theta_e^{(n)});\theta_t^{(n)}\bigr),\quad
  n=1,\dots,N,
\]
where $(\theta_e^{(n)},\theta_t^{(n)})$ denotes the $n$-th dropout realisation.  The predictive mean and total variance are
\[
  \bar Y_t(x)
  = \frac1N\sum_{n=1}^N \hat Y_t^{(n)}(x),
  \qquad
  \sigma_{\mathrm{tot}}^2(x,t)
  = \frac1N\sum_{n=1}^N\bigl(\hat Y_t^{(n)}(x)-\bar Y_t(x)\bigr)^2.
\]

\subsection{Layer-Wise Variance Decomposition}
\label{sec:decomp}

The twin-network architecture has a two-stage structure---encoder then heads---and the law of total variance gives an exact split for any such composition.  Writing $\hat Y_t(x)$ as the random output over all dropout masks:
\[
  \mathrm{Var}\bigl(\hat Y_t(x)\bigr)
  = \underbrace{\mathrm{Var}_{\theta_e}\!\Bigl(\mathbb{E}_{\theta_t}\bigl[\hat Y_t(x)\mid\theta_e\bigr]\Bigr)}_{\displaystyle \sigma_{\mathrm{enc}}^2(x,t)}
  \;+\;
  \underbrace{\mathbb{E}_{\theta_e}\!\Bigl[\mathrm{Var}_{\theta_t}\bigl(\hat Y_t(x)\mid\theta_e\bigr)\Bigr]}_{\displaystyle \sigma_{\mathrm{head}}^2(x,t)}.
\]
The first term, $\sigma_{\mathrm{enc}}^2$, measures how much the output varies due to dropout in the encoder alone.  The second, $\sigma_{\mathrm{head}}^2$, measures the average variability introduced by head dropout conditional on a fixed encoder realisation.  Both components measure sensitivity to dropout perturbations at different layers; the decomposition localises variance to architectural modules, which in turn suggests specific remedial actions (Section~\ref{sec:intro}).

\paragraph{Practical estimation.}
Computing the exact conditional expectations requires a nested Monte Carlo procedure.  We use a cheaper approximation via \emph{controlled dropout}:
\begin{itemize}
  \item \textbf{Encoder variance:} enable dropout only in the encoder, set heads to deterministic mode, and compute sample variance over $N$ passes:
  \[
    \hat\sigma_{\mathrm{enc}}^2(x,t)
    = \frac{1}{N}\sum_{n=1}^N
      \bigl(\hat Y^{(n)}_{t,\mathrm{enc}}(x) - \bar Y_{t,\mathrm{enc}}(x)\bigr)^2.
  \]
  \item \textbf{Head variance:} enable dropout only in the heads, set encoder to deterministic mode, and compute sample variance over $N$ passes:
  \[
    \hat\sigma_{\mathrm{head}}^2(x,t)
    = \frac{1}{N}\sum_{n=1}^N
      \bigl(\hat Y^{(n)}_{t,\mathrm{head}}(x) - \bar Y_{t,\mathrm{head}}(x)\bigr)^2.
  \]
\end{itemize}
This approximation is exact when encoder--head interaction terms are negligible.  We verify additivity ($\hat\sigma_{\mathrm{enc}}^2 + \hat\sigma_{\mathrm{head}}^2 \approx \sigma_{\mathrm{tot}}^2$) empirically in Section~\ref{sec:experiments}.

\subsection{Model Architecture and Training}
\begin{figure}[!htbp]
  \centering
  \includegraphics[width=0.8\linewidth]{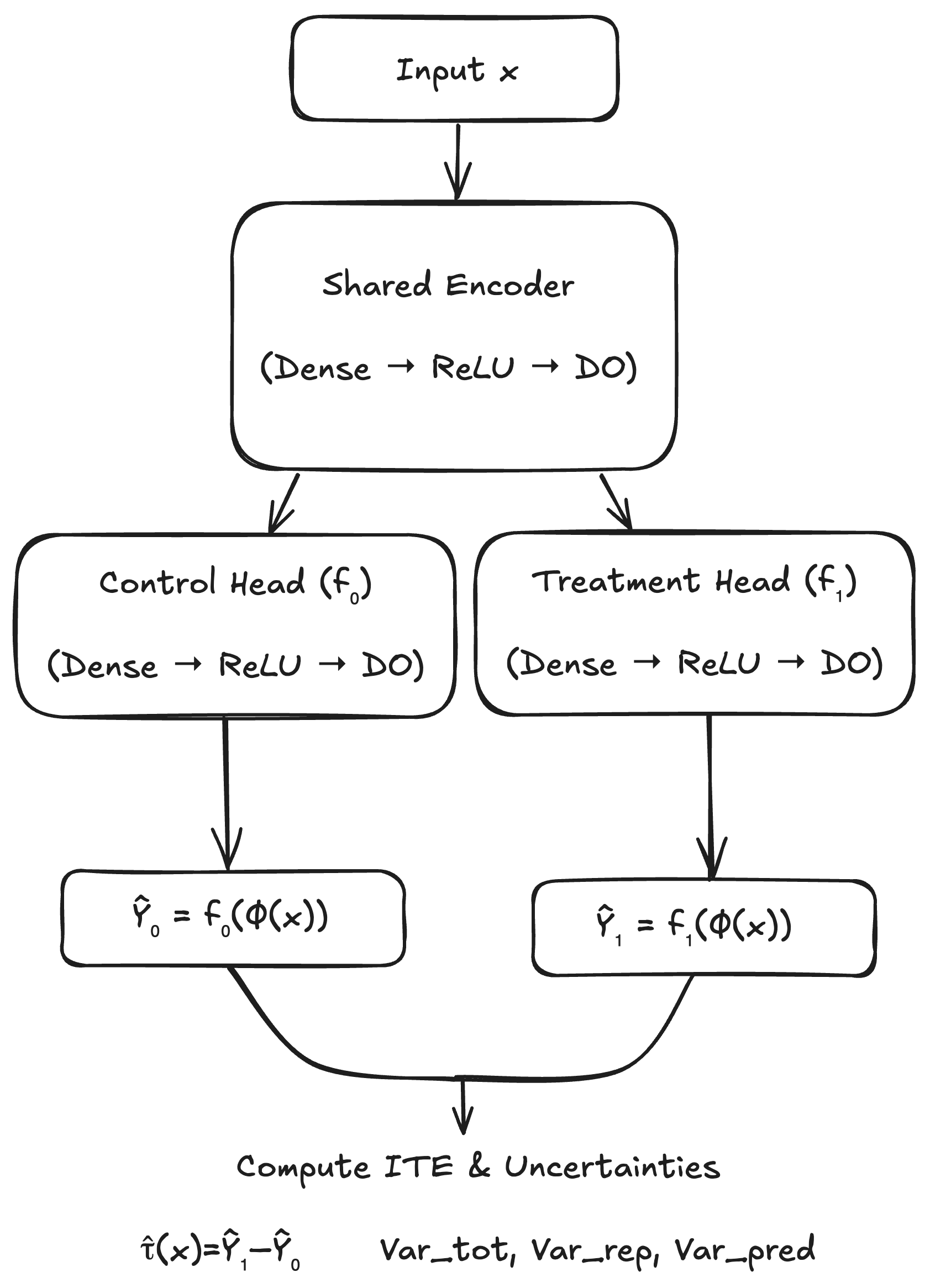}
  \caption{Twin network with layer-wise MC Dropout.  By toggling dropout independently in the encoder and heads, we obtain two variance components ($\sigma_{\mathrm{enc}}^2$, $\sigma_{\mathrm{head}}^2$) that localise uncertainty to specific architectural modules.}
  \label{fig:arch}
\end{figure}

Our implementation (Figure~\ref{fig:arch}) follows the twin-network paradigm~\cite{Shalit2017, Shi2019}.  The encoder $\Phi(x;\theta_e)$ is a multi-layer perceptron with dropout after each hidden layer.  The two outcome heads $f_0, f_1$ are identically structured MLPs, each with their own dropout layers.  Training minimises the factual mean-squared error:
\[
  \mathcal L(\theta_e,\theta_0,\theta_1)
  = \frac1n\sum_{i=1}^n
    \Bigl(\mathbb{I}[t_i=0]\,f_0(\Phi(x_i)) + \mathbb{I}[t_i=1]\,f_1(\Phi(x_i)) - y_i\Bigr)^2.
\]
Optionally, one can add an IPM-based regulariser to balance treated and control embeddings~\cite{Shalit2017}.

\subsection{Inference Procedure}
At test time, for a new input $x$ we run three sets of $N$ stochastic forward passes:
\begin{enumerate}
  \item \texttt{mode='total'}: dropout active everywhere $\;\to\;$ $\bar Y_t(x)$ and $\sigma_{\mathrm{tot}}^2(x,t)$.
  \item \texttt{mode='enc\_only'}: dropout in encoder only $\;\to\;$ $\hat\sigma_{\mathrm{enc}}^2(x,t)$.
  \item \texttt{mode='head\_only'}: dropout in heads only $\;\to\;$ $\hat\sigma_{\mathrm{head}}^2(x,t)$.
\end{enumerate}
The ITE estimate and its variance are then
\[
  \hat\tau(x)=\bar Y_1(x)-\bar Y_0(x), \qquad
  \mathrm{Var}(\hat\tau(x)) \approx \sum_{t\in\{0,1\}} \bigl[\hat\sigma_{\mathrm{enc}}^2(x,t) + \hat\sigma_{\mathrm{head}}^2(x,t)\bigr].
\]
The total cost is $3N$ forward passes per test point, compared to $N$ for standard MC Dropout.

\section{Experiments}
\label{sec:experiments}

We evaluate the layer-wise decomposition on three synthetic generators (v1--v3) and validate on a real-world twins dataset with induced multivariate sampling bias. We also compare against a standard deep ensemble baseline. All metrics are computed over 5 model seeds and 200-sample bootstraps (mean $\pm$ 95\% CI).

\subsection{Datasets and Generators}
\label{sec:datasets}
We use three synthetic data-generating processes (details in Appendix~A):
\begin{itemize}
  \item \textbf{v1 (Sin$\times$Sin)}: $Y_0 = \sin(\pi x_1 x_2) + \epsilon$,\; $\tau = 1.5 + 0.5\,x_1$.
  \item \textbf{v2 (Polynomial)}: $Y_0 = (x_1^2 + x_2^2)/2 + \epsilon$,\; $\tau = 2 + x_1 x_2$.
  \item \textbf{v3 (Sin+Linear)}: $Y_0 = \sin(x_1) + \cos(x_2) + 0.3\,x_1 x_2 + \epsilon$,\; $\tau = 1 + \sin(x_1 + x_2)$.
\end{itemize}
For each generator we apply sampling-shift and noise-shift regimes. The induced domain shift is strong in v1--v2 and mild in v3. We use an 80/20 train/test split and define OOD points as the top 20\% by a 10-NN density score.

\subsection{Metrics}
\label{sec:metrics}
We run inference in three dropout modes (Section~\ref{sec:decomp}):
\begin{itemize}
  \item \textbf{Total}: standard MC Dropout (all masks active) $\to$ $\sigma_{\mathrm{tot}}^2$.
  \item \textbf{Encoder-only}: dropout in encoder, heads deterministic $\to$ $\hat\sigma_{\mathrm{enc}}^2$.
  \item \textbf{Head-only}: dropout in heads, encoder deterministic $\to$ $\hat\sigma_{\mathrm{head}}^2$.
\end{itemize}
As a baseline we also train a 5-model deterministic ensemble and compute $\Var_{\mathrm{ens}}[\hat\tau]$.

For each test point we report Spearman's $\rho$ between each variance component and absolute ITE error $|\hat\tau - \tau|$. For the twins experiment we additionally report $\Delta\sigma^2 = \mathbb{E}[\sigma^2 \mid \mathrm{OOD}] - \mathbb{E}[\sigma^2 \mid \mathrm{ID}]$ and ROC-AUC for OOD detection.

\subsection{Implementation Details}
All models are implemented in PyTorch. We train with Adam (lr $= 10^{-3}$, weight decay $= 10^{-4}$) for 50 epochs, using dropout rate 0.2, $N = 1000$ MC samples, and batch size 128. Code and data are available at \texttt{github.com/mercury0100/TwinDrop}.

\subsection{Additivity Check}
\label{sec:additivity}

\begin{figure}[!htbp]
  \centering
  \includegraphics[width=1.0\linewidth]{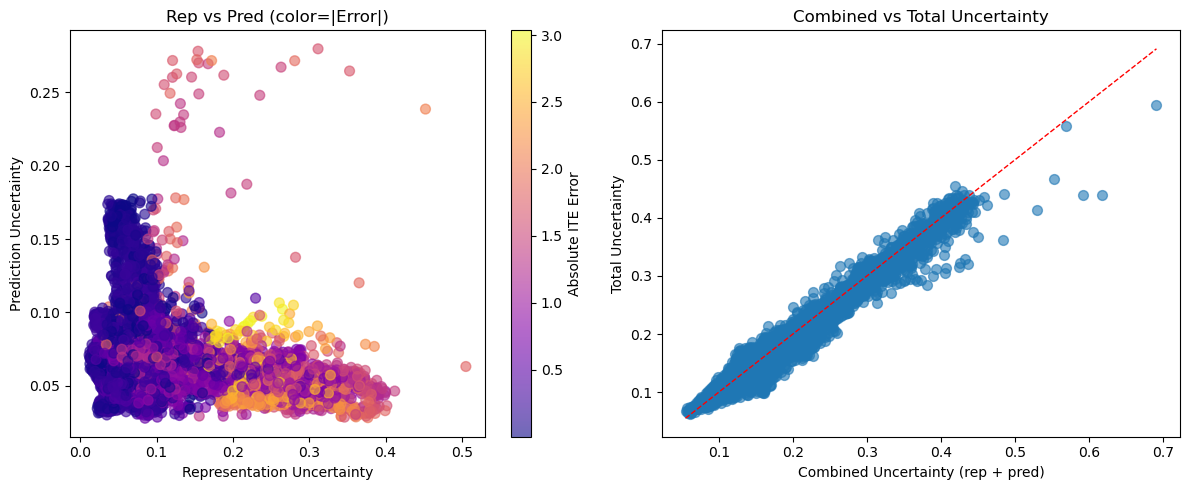}
  \caption{(Left) Variance components vs.\ ITE error on v1, colored by $|\hat\tau - \tau|$. (Right) Additivity check: $\hat\sigma_{\mathrm{enc}}^2 + \hat\sigma_{\mathrm{head}}^2$ vs.\ $\sigma_{\mathrm{tot}}^2$.}
  \label{fig:repvsperr}
\end{figure}

As noted in Section~\ref{sec:decomp}, the controlled-dropout approximation is exact only when encoder--head interaction terms are negligible. Figure~\ref{fig:repvsperr} (right) plots $\hat\sigma_{\mathrm{enc}}^2 + \hat\sigma_{\mathrm{head}}^2$ against $\sigma_{\mathrm{tot}}^2$ for all test points in v1. The close agreement ($R^2 > 0.99$) confirms that the approximation is adequate for these architectures and that the two components account for essentially all of the total dropout variance.

\subsection{Spatial Distribution of Variance Components}
\label{sec:demo}

\begin{figure}[!htbp]
  \centering
  \includegraphics[width=1.0\linewidth]{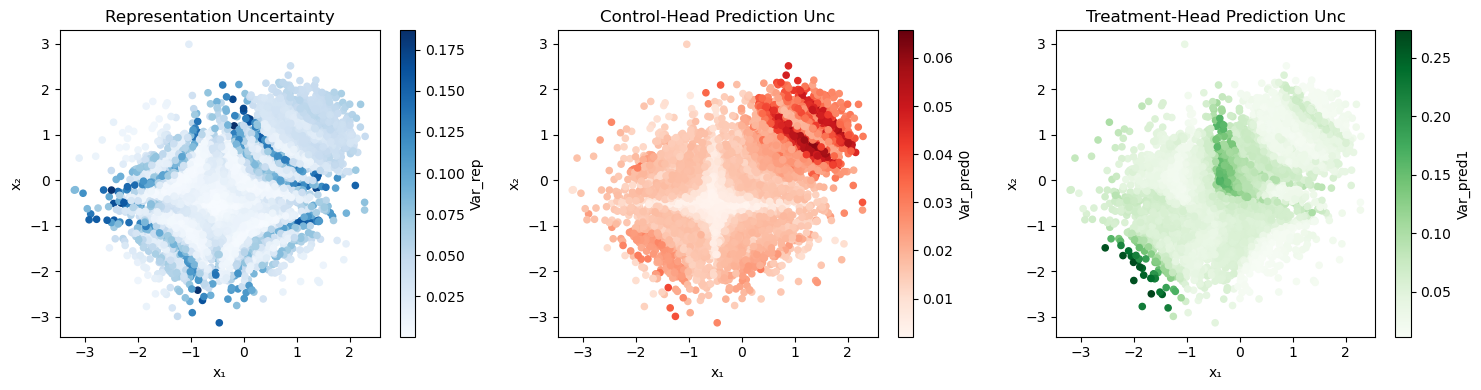}
  \caption{Variance components on v1: encoder $\hat\sigma_{\mathrm{enc}}^2$ (blue), control-head $\hat\sigma_{\mathrm{head},0}^2$ (red), treatment-head $\hat\sigma_{\mathrm{head},1}^2$ (green).}
  \label{fig:demo-v1}
\end{figure}

Figure~\ref{fig:demo-v1} shows the two variance components over the $(x_1, x_2)$ plane for v1 under sampling- and noise-shift. The encoder component $\hat\sigma_{\mathrm{enc}}^2$ is elevated in regions where training data is sparse (the shifted region), while the head components are more uniformly distributed. This spatial pattern is consistent with the encoder component tracking covariate coverage, though we emphasize that this is an empirical observation on this particular DGP, not a guaranteed property of the decomposition.

\subsection{Error Prediction: Which Component Tracks ITE Error?}
\label{sec:error_prediction}

Table~\ref{tbl:error_prediction} reports Spearman's $\rho$ between each variance component and $|\hat\tau - \tau|$ across the three generators.

\begin{table}[!htbp]
\centering
\small
\caption{Spearman's $\rho(\sigma^2, |\hat\tau - \tau|)$ by generator (mean $\pm$ 95\% CI).}
\label{tbl:error_prediction}
\begin{tabular}{lccc}
\toprule
Metric & v1 & v2 & v3 \\
\midrule
$\rho_{\mathrm{enc}}$    & $0.5305 \pm 0.0166$ & $0.3443 \pm 0.0219$ & $0.3933 \pm 0.0215$ \\
$\rho_{\mathrm{head}}$   & $-0.0506 \pm 0.0244$ & $0.0831 \pm 0.0246$ & $0.3250 \pm 0.0229$ \\
$\rho_{\mathrm{tot}}$    & $0.4493 \pm 0.0209$ & $0.2110 \pm 0.0224$ & $0.4201 \pm 0.0221$ \\
$\rho_{\mathrm{head},0}$ & $-0.0017 \pm 0.0276$ & $0.0435 \pm 0.0295$ & $0.0828 \pm 0.0317$ \\
$\rho_{\mathrm{head},1}$ & $-0.3305 \pm 0.0395$ & $0.1172 \pm 0.0480$ & $0.2884 \pm 0.0438$ \\
\bottomrule
\end{tabular}
\end{table}

The pattern across generators is consistent: in the strong-shift regimes (v1, v2), the encoder component is the dominant error predictor ($\rho_{\mathrm{enc}} = 0.53$ and $0.34$), while the head component is near zero or slightly negative. Under mild shift (v3), both components carry signal, but the encoder component still leads ($\rho_{\mathrm{enc}} = 0.39$ vs.\ $\rho_{\mathrm{head}} = 0.33$). Total variance, which averages both signals, is a weaker predictor than the encoder component alone whenever shift is the primary source of error.

These correlations are moderate, not strong. We view them as evidence that the decomposition provides a \emph{useful ranking signal}---points with high encoder variance tend to have higher error---rather than a precise error estimate.

\begin{figure}[!htbp]
  \centering
  \includegraphics[width=0.48\linewidth]{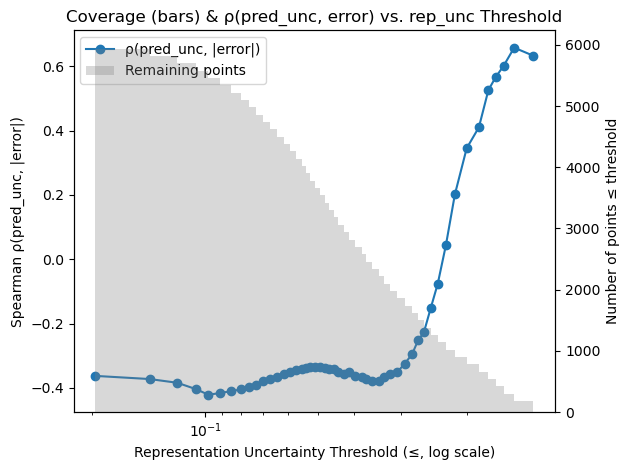}
  \hfill
  \includegraphics[width=0.48\linewidth]{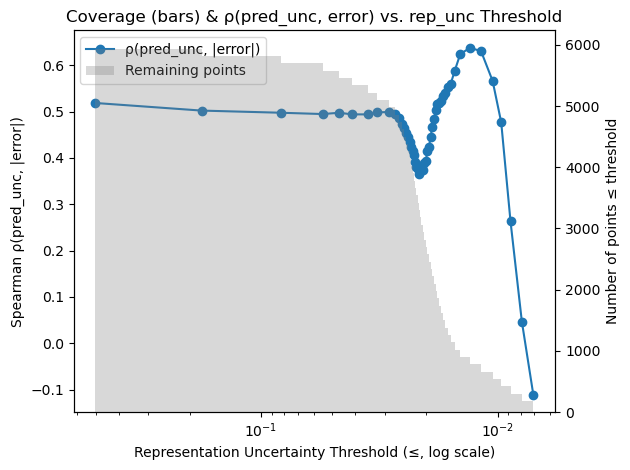}
  \caption{$\rho(\hat\sigma_{\mathrm{head}}^2, |\hat\tau - \tau|)$ vs.\ maximum allowed $\hat\sigma_{\mathrm{enc}}^2$ for v1 (left) and v3 (right). Once high-encoder-variance points are excluded, the head component becomes a strong predictor of ITE error.}
  \label{fig:rho_vs_threshold}
\end{figure}

\paragraph{Conditional informativeness of the head component.}
The head component appears uninformative in Table~\ref{tbl:error_prediction}, but this is partly because it is swamped by the encoder signal on OOD points. To test whether it carries independent information, we sweep a maximum-$\hat\sigma_{\mathrm{enc}}^2$ threshold and recompute $\rho(\hat\sigma_{\mathrm{head}}^2, |\hat\tau - \tau|)$ on the remaining points. Figure~\ref{fig:rho_vs_threshold} shows that as we filter out high-encoder-variance points, the head component becomes a strong error predictor ($\rho_{\mathrm{head}} > 0.5$) in both v1 and v3. This suggests a two-stage diagnostic: first check encoder variance to flag potential OOD points, then use head variance to rank uncertainty among the in-distribution remainder.

\subsection{Calibration}
\label{sec:calibration}

\begin{figure}[!htbp]
  \centering
  \includegraphics[width=0.48\linewidth]{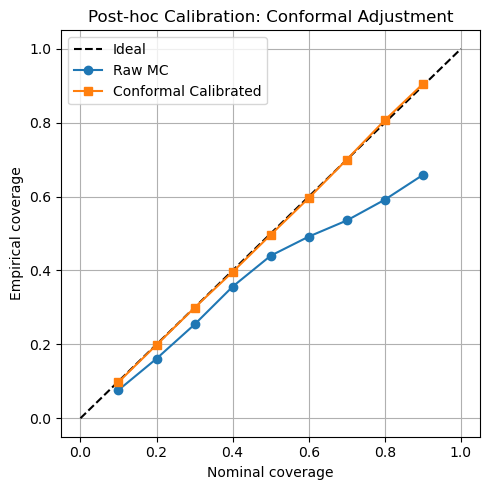}
  \includegraphics[width=0.48\linewidth]{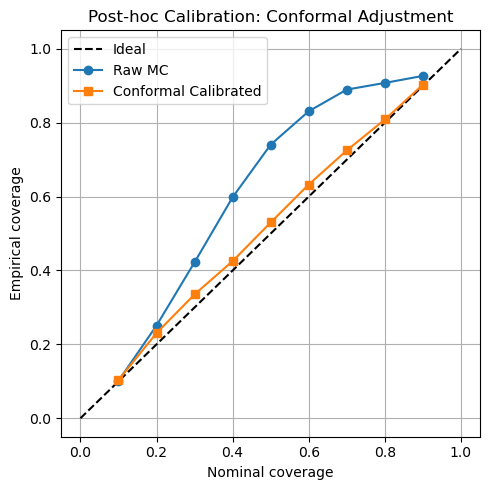}
  \caption{Reliability diagrams for v1 (left) and v3 (right): raw MC Dropout (blue) vs.\ conformal-adjusted (orange). Raw ECE: 0.104 (v1), 0.130 (v3); post-hoc ECE: 0.005 (v1), 0.022 (v3).}
  \label{fig:calibration_post}
\end{figure}

We assess interval calibration before and after post-hoc conformal adjustment using a held-out calibration fold. Raw MC Dropout intervals exhibit nontrivial miscalibration (ECE $= 0.104$ on v1, $0.130$ on v3). After conformal adjustment, ECE falls to $0.005$ (v1) and $0.022$ (v3). Figure~\ref{fig:calibration_post} shows the reliability curves. The decomposition itself does not change calibration---the same conformal correction applies to $\sigma_{\mathrm{tot}}^2$---but well-calibrated total intervals are a prerequisite for the component-level diagnostics to be meaningful.

\subsection{Comparison with Deep Ensembles}
\label{sec:ensemble}

Table~\ref{tbl:ensemble} compares total-variance error correlation for MC Dropout against a 5-model deterministic ensemble.

\begin{table}[!htbp]
\centering
\small
\caption{Spearman's $\rho(\sigma_{\mathrm{tot}}^2, |\hat\tau - \tau|)$: MC Dropout vs.\ 5-model ensemble.}
\label{tbl:ensemble}
\begin{tabular}{lcc}
\toprule
Generator & MC Dropout $\rho_{\mathrm{tot}}$ & Ensemble $\rho$ \\
\midrule
v1 (strong shift)   & $0.4493 \pm 0.0209$ & $0.834$ \\
v2 (moderate shift)  & $0.2110 \pm 0.0224$ & $0.036$ \\
v3 (mild shift)      & $0.4201 \pm 0.0221$ & $0.040$ \\
\bottomrule
\end{tabular}
\end{table}

The results highlight complementary failure modes rather than a clear winner. Under strong shift (v1), the ensemble produces much higher error correlation ($\rho = 0.83$), likely because the five models learn genuinely different functions in the data-sparse shifted region, producing meaningful inter-model disagreement. Under moderate and mild shift (v2, v3), the ensemble collapses to near-zero correlation ($\rho \approx 0.04$), suggesting that the individual models converge to essentially the same solution and inter-model variance no longer tracks error. MC Dropout, by contrast, maintains a moderate signal across all regimes ($\rho \approx 0.21$--$0.45$) because dropout continues to inject stochasticity regardless of model convergence.

Neither method dominates. The ensemble is superior when shift is strong enough to induce genuine model disagreement; MC Dropout is more robust when models converge. A practical implication is that combining the two---for instance, applying our layer-wise decomposition within each ensemble member---could yield both the diversity benefits of ensembles and the localization benefits of the decomposition, though we leave this to future work.

\subsection{Real-World Validation: Twins with Induced Shift}
\label{sec:twins_multivariate}

As a sanity check on the synthetic findings, we apply the decomposition to the same-sex twins cohort. We induce a multivariate sampling bias along the first principal component and split test points into OOD (top 20\% by 10-NN density) vs.\ ID. Table~\ref{tbl:twins_shift} summarizes the results.

\begin{table}[!htbp]
\centering
\small
\caption{Twins cohort: variance component behavior before and after induced sampling bias (mean $\pm$ 95\% CI). OOD defined as top 20\% by 10-NN density.}
\label{tbl:twins_shift}
\begin{tabular}{lccc|ccc}
\toprule
 & \multicolumn{3}{c}{\bfseries No Bias} & \multicolumn{3}{c}{\bfseries With Bias} \\
Metric & Mean & 2.5\% & 97.5\% & Mean & 2.5\% & 97.5\% \\
\midrule
$\Delta\sigma_{\mathrm{enc}}^2$    & 0.00366 & 0.00259 & 0.00469 & 0.00385 & 0.00088 & 0.00696 \\
$\Delta\sigma_{\mathrm{head}}^2$   & 0.00020 & 0.00005 & 0.00036 & 0.00003 & $-$0.00035 & 0.00042 \\
$\Delta\sigma_{\mathrm{tot}}^2$    & 0.00370 & 0.00254 & 0.00477 & 0.00348 & 0.00035 & 0.00664 \\
\midrule
$\rho(\sigma_{\mathrm{enc}}^2, |e|)$    & 0.761 & 0.743 & 0.783 & 0.890 & 0.864 & 0.917 \\
$\rho(\sigma_{\mathrm{head}}^2, |e|)$   & 0.810 & 0.794 & 0.828 & 0.850 & 0.822 & 0.876 \\
$\rho(\sigma_{\mathrm{tot}}^2, |e|)$    & 0.779 & 0.762 & 0.797 & 0.899 & 0.871 & 0.928 \\
$\rho(\sigma_{\mathrm{head},0}^2, |e|)$ & 0.785 & 0.760 & 0.811 & 0.739 & 0.691 & 0.783 \\
$\rho(\sigma_{\mathrm{head},1}^2, |e|)$ & 0.432 & 0.397 & 0.470 & 0.648 & 0.591 & 0.696 \\
\bottomrule
\end{tabular}
\end{table}

The key observations are consistent with the synthetic experiments. First, the encoder component spikes on OOD points ($\Delta\sigma_{\mathrm{enc}}^2 \approx 0.004$) while the head component remains flat ($\Delta\sigma_{\mathrm{head}}^2 \approx 0$), confirming that the decomposition separates shift-sensitive and shift-insensitive signals even on real data. Second, under induced bias the encoder component becomes the strongest single error predictor ($\rho_{\mathrm{enc}} = 0.89$), though total variance is comparable ($\rho_{\mathrm{tot}} = 0.90$) because the encoder component dominates the sum.

We note two important caveats. The shift here is artificially induced, so the ``OOD'' region is constructed rather than naturally occurring. And the Spearman correlations on twins are substantially higher across the board ($\rho > 0.75$) than on the synthetic data, likely because the twins dataset has lower intrinsic noise. We present this experiment as evidence that the decomposition \emph{transfers} to real data, not as a standalone validation.

We deliberately omit ROC-AUC for OOD detection from the main narrative: the values (0.54--0.62) are too close to chance to support a claim that the decomposition is a reliable OOD detector. The decomposition's value lies in error prediction and diagnostic localization, not in binary OOD classification.

\section{Discussion}
\label{sec:discussion}

The experiments in Section~\ref{sec:experiments} tell a consistent story: toggling dropout at different architectural modules produces variance components that behave differently under distributional shift, and this difference carries practical diagnostic value. Here we summarize the main findings, address limitations honestly, and outline directions for future work.

\paragraph{The encoder component tracks shift-induced error.}
Across all three synthetic generators, the encoder component $\hat\sigma_{\mathrm{enc}}^2$ is the strongest single predictor of ITE error whenever covariate shift is the dominant source of failure (v1: $\rho_{\mathrm{enc}} = 0.53$; v2: $\rho_{\mathrm{enc}} = 0.34$). Under mild shift (v3), the advantage narrows ($\rho_{\mathrm{enc}} = 0.39$ vs.\ $\rho_{\mathrm{tot}} = 0.42$), which is expected: when shift is not the primary error source, localizing variance to the encoder buys less. The twins experiment reinforces this pattern---under induced bias, $\rho_{\mathrm{enc}}$ rises to $0.89$---though the artificially constructed shift limits how much weight we place on this result.

\paragraph{The head component carries conditional signal.}
In isolation, the head component $\hat\sigma_{\mathrm{head}}^2$ is a weak error predictor (Table~\ref{tbl:error_prediction}). But the threshold-sweep analysis (Figure~\ref{fig:rho_vs_threshold}) reveals that this is because it is swamped by encoder-level variation on OOD points. Once high-$\hat\sigma_{\mathrm{enc}}^2$ points are excluded, head variance becomes a strong predictor ($\rho_{\mathrm{head}} > 0.5$). This suggests a natural two-stage workflow: use the encoder component to flag regions where the model may be extrapolating, then use the head component to rank residual uncertainty among in-distribution points.

\paragraph{Ensembles and dropout have complementary strengths.}
The ensemble comparison (Table~\ref{tbl:ensemble}) does not show a clear winner. Under strong shift (v1), a 5-model ensemble produces much higher error correlation ($\rho = 0.83$) than MC Dropout ($\rho = 0.45$), because the models learn genuinely different functions in the data-sparse region. Under moderate and mild shift (v2, v3), the ensemble collapses ($\rho \approx 0.04$) as the models converge to near-identical solutions. MC Dropout maintains a moderate signal throughout because dropout injects stochasticity regardless of convergence. The practical takeaway is that neither method dominates, and combining them---applying the layer-wise decomposition within each ensemble member---is a natural extension.

\paragraph{Limitations.}
We highlight several limitations that qualify the conclusions above.

First, the synthetic generators are low-dimensional (2--3 covariates) with known treatment-effect functions. Whether the patterns we observe---especially the dominance of encoder variance under shift---hold in high-dimensional settings with complex, unknown treatment heterogeneity remains to be tested.

Second, the twins experiment uses artificially induced shift rather than naturally occurring distributional differences. The high Spearman correlations ($\rho > 0.75$ across the board) likely reflect the low intrinsic noise of the twins dataset rather than a strong property of the method. A more convincing validation would involve naturally shifted subgroups or an entirely separate observational dataset.

Third, MC Dropout is a rough posterior approximation. The variance it produces depends on the dropout rate, the network architecture, and the training procedure. Our results are specific to the architectures and hyperparameters tested; sensitivity to these choices deserves further study.

\paragraph{Future work.}
The most immediate extension is a \emph{targeted data collection} experiment: when $\hat\sigma_{\mathrm{enc}}^2$ is high in a region, collect additional training data from that region and verify that encoder variance drops and ITE error improves; when $\hat\sigma_{\mathrm{head}}^2$ is high, verify that additional outcome replicates help instead. This would provide direct evidence that the decomposition is actionable, not just descriptive.

Other directions include extending the decomposition to richer posterior approximations (e.g., variational inference or SWAG), applying it within ensemble members to combine the benefits of both approaches, testing on larger observational datasets with natural subgroup shifts, and exploring whether the same layer-wise principle applies to other modular architectures beyond twin networks (e.g., multi-task models, encoder--decoder architectures in NLP).

\section{Conclusion}
\label{sec:conclusion}

We have presented a layer-wise variance decomposition for deep twin-network models that splits total MC Dropout variance into an encoder component and a head component via the law of total variance. The decomposition is cheap (two additional sets of forward passes), exact in principle (with an approximation we validate empirically), and produces two signals with distinct diagnostic behaviour: the encoder component tracks covariate-shift-induced error, while the head component becomes informative once encoder-level uncertainty is controlled.

Our decomposition provides is architectural localisation: a practitioner can see \emph{where} in the model uncertainty concentrates, and this points toward specific remedial actions (diversify covariate coverage vs.\ add outcome data). Across three synthetic generators and a real-world twins cohort, this localisation is empirically useful, producing error-prediction signals that are as strong as or stronger than monolithic total variance.

The correlations we report are moderate on synthetic data ($\rho \approx 0.3$--$0.5$) and stronger on twins ($\rho \approx 0.9$), and we have been transparent about the limitations of both settings. We view this work as a proof of concept for a diagnostic tool, not a solved problem. The most important next step is demonstrating that the decomposition leads to better decisions---targeted data collection, selective prediction, or adaptive model updating---rather than just better uncertainty descriptions.

Code and data are available at \url{https://github.com/mercury0100/TwinDrop}.

\bibliographystyle{unsrt}
\bibliography{references}

\end{document}